\documentclass[sigconf]{acmart}
\DeclareMathAlphabet      {\mathbf}{OT1}{cmr}{bx}{n}

\usepackage{CJKutf8}
\usepackage[linesnumbered,ruled,vlined]{algorithm2e}
\usepackage{amsmath,amsfonts}

\DeclareMathOperator*{\argmax}{argmax}

\AtBeginDocument{
  \providecommand\BibTeX{{
    \normalfont B\kern-0.5em{OT1/cmr/sb/n\scshape i\kern-0.25em b}\kern-0.8em\TeX}}}

\setcopyright{iw3c2w3}
\begin{document}

\copyrightyear{2021}
\acmYear{2021}
\acmConference[WWW '21]{Proceedings of the Web Conference 2021}{April 19--23, 2021}{Ljubljana, Slovenia}
\acmBooktitle{Proceedings of the Web Conference 2021 (WWW '21), April 19--23, 2021, Ljubljana, Slovenia}
\acmPrice{}
\acmDOI{10.1145/3442381.3449931}
\acmISBN{978-1-4503-8312-7/21/04}
\settopmatter{printacmref=true}

\title{\textit{Verdi}: Quality Estimation and Error Detection for Bilingual Corpora}

\author{Mingjun Zhao}
\authornote{Both authors contributed equally to this research.}
\affiliation{
  \institution{University of Alberta}
  \city{Edmonton}
  \country{Canada}}
\email{zhao2@ualberta.ca}

\author{Haijiang Wu}
\authornotemark[1]
\affiliation{
  \institution{Tencent}
  \city{Shenzhen}
  \country{China}}
\email{harywu@tencent.com}

\author{Di Niu}
\affiliation{
  \institution{University of Alberta}  
  \city{Edmonton}
  \country{Canada}}
\email{dniu@ualberta.ca}

\author{Zixuan Wang}
\affiliation{
  \institution{Tencent}
  \city{Shenzhen}
  \country{China}}
\email{zackiewang@tencent.com}

\author{Xiaoli Wang}
\affiliation{
  \institution{Tencent}
  \city{Shenzhen}
  \country{China}}
\email{evexlwang@tencent.com}

\begin{abstract}
Translation Quality Estimation is critical to reducing post-editing efforts in machine translation and to cross-lingual corpus cleaning.
As a research problem, quality estimation (QE) aims to directly estimate the quality of translation in a given pair of source and target sentences, and highlight the words that need corrections, without referencing to golden translations.
In this paper, we propose 
\textit{Verdi}, a novel framework
for word-level and sentence-level post-editing effort estimation for bilingual corpora.  
\textit{Verdi} adopts two word predictors to enable diverse features to be extracted from a pair of sentences for subsequent quality estimation, including a transformer-based neural machine translation (NMT) model and a pre-trained cross-lingual language model (XLM). 
We exploit the symmetric nature of bilingual corpora and apply model-level dual learning in the NMT predictor, which handles a primal task and a dual task simultaneously with weight sharing, leading to stronger context prediction ability than single-direction NMT models.
By taking advantage of the dual learning scheme, we further design a novel feature to directly encode the translated target information without relying on the source context. 
Extensive experiments conducted on WMT20 QE tasks demonstrate that our method beats the winner of the competition and outperforms other baseline methods by a great margin.
We further use the sentence-level scores provided by \textit{Verdi} to clean a parallel corpus and observe benefits on both model performance and training efficiency.
\end{abstract}



\maketitle

\section{Introduction}

Neural Machine Translation (NMT) has achieved great success in generating accurate translations with a single neural network in an end-to-end fashion. 
Although NMT eliminates the need for hand-crafted features and rules, as compared to traditional statistical machine translation methods,
yet a well-performing NMT model is largely dependent on large amounts of high-quality training data. However,  acquiring accurate translation pairs is a costly task requiring cross-lingual expertise.
Common practices of creating a large translation dataset include leveraging the parallel corpus crawled from web, e,g., Wikipedia, and/or relying on existing machine translators to generate sentence pairs. 
However, the web-crawled or machine-generated corpora usually do not have any guarantee in quality. They contain a high degree of noise \cite{wang2018denoising} and demand further editing or filtering to correct the errors, which is still a highly expensive procedure. As a result, the ability of automatically estimating the translation quality of sentence pairs becomes an urgent need.

Recently, Quality Estimation (QE) has become of growing importance in the field of machine translation and cross-lingual corpus filtering, which aims to directly assess the quality of a source-target translation pair, without referencing to golden human translations, with typically two goals: word-level and sentence-level post-editing effort estimations \cite{fan2019bilingual,kim2017predictor,kepler2019openkiwi}.
The goal of word-level QE is to predict a tag for each token in the translated sentence to detect the mistranslated words and a tag for each gap between two tokens to detect missing words. 
On the other hand, in sentence-level QE tasks, a commonly  used measurement metric is the
Human Translation Error Rate (HTER) score \cite{snover2006study}, which is the ratio between the number of edits needed and the reference translation length.
Table \ref{tab:example} gives a data example from the WMT20 English-Chinese QE dataset, where the accurate translations are shown in green and the bad translations are marked in red.

Traditional studies on quality estimation are based on handcrafted features  \cite{ueffing2007word,specia-etal-2013-quest,kozlova2016ysda} or feature selection \cite{shah2015bayesian,beck2013shef}.
Deep learning methods have also been applied to quality estimation \cite{kreutzer2015quality,martins2017pushing}. 
The recent Predictor-Estimator framework \cite{kim2017predictor} for QE has drawn much attention, which consists of an RNN word prediction model firstly trained on a large amount of parallel corpora for feature extractions, and an estimator that is subsequently trained on a smaller amount of sentence pairs with QE labels and provides quality evaluation based on the extracted features.
\citeauthor{fan2019bilingual} \cite{fan2019bilingual} base their work on the predictor-estimator framework and further propose to substitute the RNN predictor with a bidirectional transformer language model.

In this paper, we propose \textit{Verdi}, a novel translation quality estimation and error detection method, which falls into the Predictor-Estimator framework, yet leveraging model-level dual learning, mixture models and cross-lingual language models to extract a range of diverse features to boost performance on both sentence-level and word-level QE tasks. Our framework utilizes two word predictors for feature extraction, including an NMT model and a pre-trained XLM model. 

Taking advantage of the symmetric nature of parallel corpora, our NMT word predictor is a transformer \cite{vaswani2017attention} reformulated by the model-level dual learning framework \cite{xia2018model},
 which behaves as a regularization method that ties the model parameters that are playing similar roles in two dual tasks: the forward and the backward translation tasks.
In particular, we reformulate Transformer into two unified conditional self-attentive encoders that can handle the two dual tasks simultaneously with the same set of parameters, where one conditional encoder works as both the encoder in the source-to-target translation and the decoder in the target-to-source translation. The opposite applies to the other encoder.
By training on two dual tasks with double the amount of training data and weight sharing, our NMT predictor is capable of outputting more accurate translations and producing more robust representations for the sentence pair.
Additionally, we have designed a novel Dual Model Feature for quality estimation based on the model-level dual learning framework.
The dual model feature models the internal correlations in the translated sentence without relying on the source context information, and can effectively improve the performance of quality estimation.

Furthermore, we utilize the mixture models technique in our framework to train the NMT predictor, where multiple expert models are trained and combined.
In particular, each expert model is assigned with a different start-of-sentence token, and is trained with the EM algorithm \cite{mclachlan2007algorithm}, so that different expert models can specialize on different translation styles.
Therefore, our NMT predictor is capable of producing diverse translation outputs, thus accommodating the one-to-many setup of machine translation.

The second predictor we utilize is a pre-trained cross-lingual language model (XLM) \cite{conneau2019cross} to further improve the feature extraction ability. 
Unlike monolingual language models such as BERT \cite{devlin2018bert}, XLM models are pre-trained on parallel corpora and construct cross-lingual representations on text, which suit better to the translation quality estimation task.
In contrast to the single predictor design in most prior work, e.g., \cite{fan2019bilingual,kim2017predictor,kepler2019openkiwi,kepler2019unbabel}, our two predictors are capable of modeling the sentence pair from different perspectives, leading to a more compact representation, hence improving the estimation results.

Extensive experiments conducted on the WMT20 QE competition dataset demonstrate that our method beats the winning system of the competition and outperforms the baseline methods by a great margin on both word-level and sentence-level QE tasks.
We have also designed and conducted experiments of exploiting the QE estimation results as the filtering criterion for cleaning a bilingual corpus. 
The outcome shows that compared with using all unfiltered samples, training on the high-quality corpus filtered by our method can yield a better NMT system, as well as save training costs.
Our source code is available at \url{https://github.com/Simpleple/Verdi}.

\begin{table}[tb]
  \centering
  \begin{tabular}{ll}
    \toprule
    Source & Many butterflies flutter among the flowers \\
    & and grasses .\\
    MT\textit{-back} & \textcolor{ACMGreen}{Many butterflies} \textcolor{ACMRed}{float in} \textcolor{ACMGreen}{the flowers}\\
    & \textcolor{ACMGreen}{and grasses .}\\
    MT & \textcolor{ACMGreen}{\begin{CJK*}{UTF8}{gbsn}许多\end{CJK*}  \begin{CJK*}{UTF8}{gbsn}蝴蝶\end{CJK*}  \begin{CJK*}{UTF8}{gbsn}在\end{CJK*}  \begin{CJK*}{UTF8}{gbsn}花草\end{CJK*}}  \textcolor{ACMRed}{\begin{CJK*}{UTF8}{gbsn}间\end{CJK*}  \begin{CJK*}{UTF8}{gbsn}飘动\end{CJK*}  .}\\
    PE (reference) & \begin{CJK*}{UTF8}{gbsn}许多\end{CJK*}  \begin{CJK*}{UTF8}{gbsn}蝴蝶\end{CJK*}  \begin{CJK*}{UTF8}{gbsn}在\end{CJK*}  \begin{CJK*}{UTF8}{gbsn}花草\end{CJK*}  \begin{CJK*}{UTF8}{gbsn}丛中\end{CJK*}  \begin{CJK*}{UTF8}{gbsn}飞舞\end{CJK*}  \begin{CJK*}{UTF8}{gbsn}。\end{CJK*}\\
    Word tags & \textbf{\textcolor{ACMGreen}{OK OK OK OK} \textcolor{ACMRed}{BAD BAD BAD}}\\
    Gap Tags & \textbf{\textcolor{ACMGreen}{OK OK OK OK OK OK OK OK}}\\
    HTER Score & 0.4286 \\
    \bottomrule
  \end{tabular}
  \caption{An example from the WMT20 QE dataset, consisting of the English source sentence, the translated sentence in Chinese (MT) and the back translation in English for demonstration, the reference translation post-edited from MT (PE), and the QE labels including word tags, gap tags and the HTER score.}
  \label{tab:example}
  \vspace{0mm}
\end{table}

\section{Problem Formulation}

In this section, we provide a brief introduction of the background and the problem formulation of the quality estimation for machine translation.

The task of machine translation is to map a sentence $x$ in the source language to its translation $y$ in the target language. 
Modern neural machine translation models adopt an encoder-decoder architecture where an encoder $p(\mathbf{h}|x)$ maps the source sentence $x$ into a latent representation $\mathbf{h}$ and a decoder $p(y|\mathbf{h})=\sum_{i=1}^L p(y_i|y_{<i},h)$ decodes the target sentence $y$ based on the latent vector $\mathbf{h}$ and its previous outputs $\{y_1,...,y_{i-1}\}$, where L is the length of the target sentence.

Quality estimation aims to provide an evaluation on the quality of the translation pair $(x,y)$, either in the detailed word level or in the overall sentence level. 
In word-level quality estimation, each word in the translated sentence is given a binary tag $t$ of either ``OK'' or ``BAD'' representing the correctness of the translation.
Similarly, for each gap in the target sentence, a tag is also provided to indicate whether one or more missing words should be added at the gap position.
Therefore, word-level quality estimation can be solved with a typical sequential labeling model $p(\mathbf{t}|x,y)$ predicting a binary vector $\mathbf{t}$ for all the words and gaps given the source sentence $x$ and the translation output $y$.

Human Translation Error Rate (HTER) is defined as the ratio between the number of post-editing operations needed to modify the translation output to the golden reference and the number of tokens in the translation sentence. 
The sentence-level quality estimation task aims to predict the HTER score $l$ for a translation sentence pair and a regression model $p(l|x,y)$ can be learned to solve it.

\section{Model Description}

\begin{figure}[tb]
\centering
\includegraphics[width=2.2in]{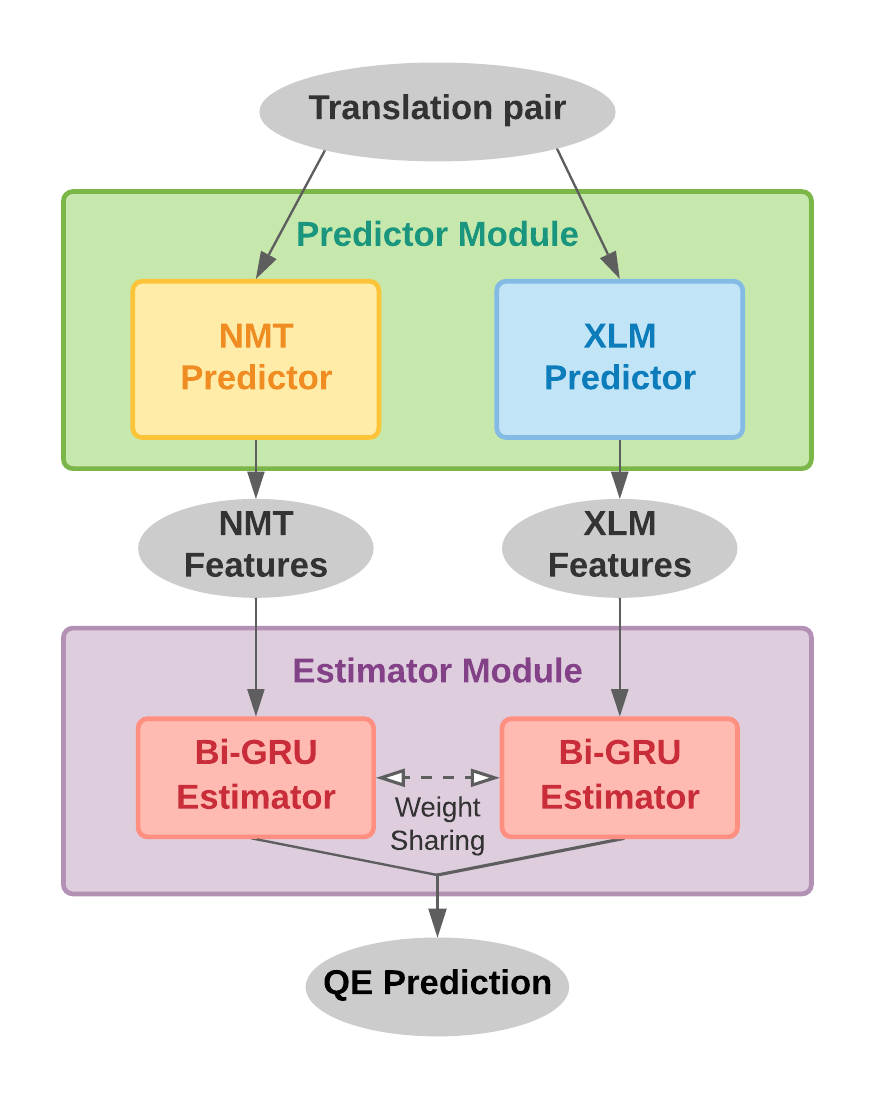}
\caption{Illustration of the overall architecture of our proposed framework. It contains a predictor module consisting of an NMT predictor and an XLM predictor for feature extraction, and an estimator module to evaluate the translation quality.}
\label{fig:system}
\vspace{0mm} 
\end{figure}

In this section, we introduce our proposed framework in detail.
Similar to \cite{kim2017predictor,fan2019bilingual}, our \textit{Verdi} framework for quality estimation is based on the predictor-estimator architecture, where a predictor module is first trained on a parallel corpus and is responsible for extracting features, and an estimator module is trained on QE datasets to evaluate the translation quality by taking the extracted features as input. 
Fig.\ref{fig:system} illustrates the overall architecture of our framework which mainly consists of three components: the NMT predictor, the XLM predictor, and the QE estimator.

The NMT predictor is a neural machine translation model based on transformer architecture, that automatically transforms sentences in the source language to their translations in the target language. 
For the purpose of sufficiently exploiting the symmetric structure of machine translation, we propose to adopt model-level dual learning on the NMT predictor, and train it on both directions of translation.
Additionally, we apply the method of mixture models \cite{shen2019mixture} for diverse machine translation during the training of the NMT predictor, in order to increase the diversity of the translation outputs and derive more robust features for evaluation.

The XLM predictor utilizes a pre-trained XLM model that takes in both sentences in a translation pair as input, and produces cross-lingual feature representations.

For the design of QE features, we first borrow the idea of quality estimation feature vectors (QEFVs) and mis-matching features from \cite{kim2017predictor,fan2019bilingual}, and propose a novel feature named Dual Model Feature.
QEFVs can be seen as the approximated transferred knowledge from word prediction to quality estimation, and mis-matching features measure the difference between the model predictions and the given translated sentence.
Apart from the above features, our proposed dual model feature takes the advantage of model-level dual learning, and focuses on modeling the translation information without source contexts involved.

Because the QE datasets are usually small with a few thousand samples, it is very likely to overfit to the dataset.
Thus, we incorporate bidirectional GRU estimators with weight sharing to transform the extracted NMT and XLM features into the final quality prediction, to alleviate the overfitting problem.

\begin{figure}[tb]
\centering
\includegraphics[width=2.8in]{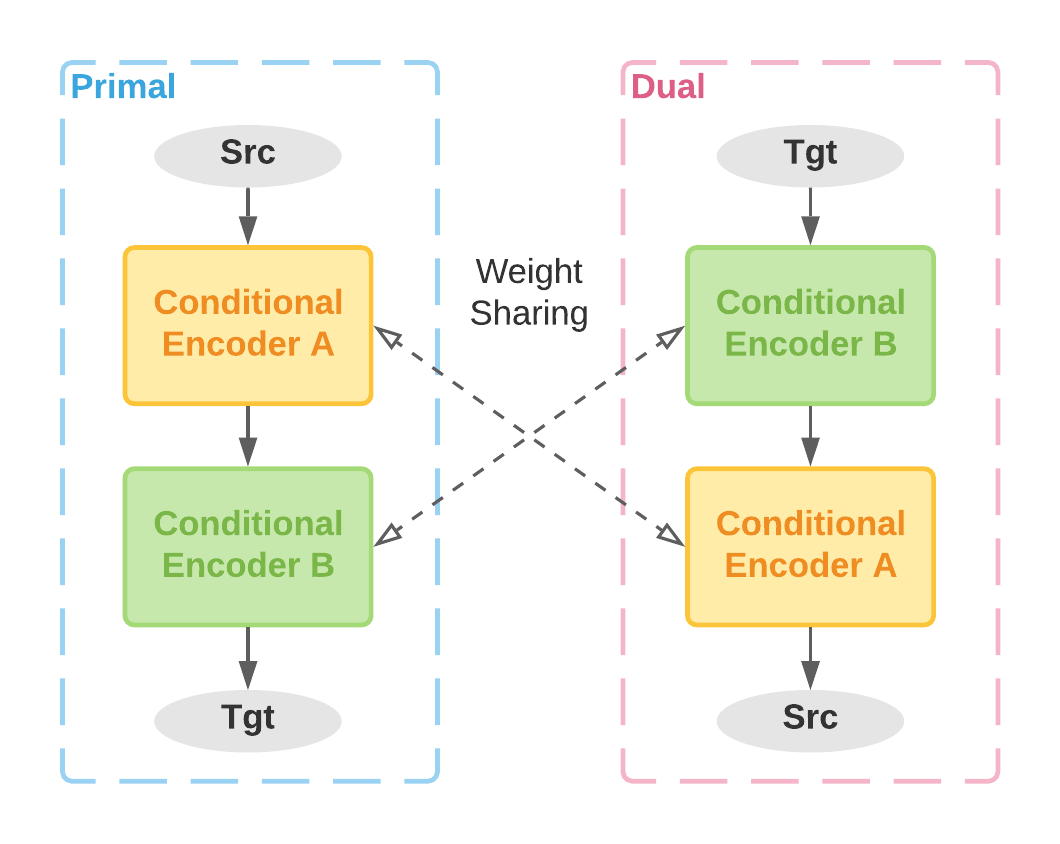}
\caption{Illustration of the training scheme of the NMT predictor with model-level dual learning.}
\label{fig:dual}
\vspace{0mm} 
\end{figure}

\subsection{NMT Predictor}
\subsubsection{Model-level Dual Learning}

Transformer \cite{vaswani2017attention} is the current state-of-the-art model in most machine translation tasks, which incorporates the encoder-decoder architecture and is solely built on the multi-headed  attention mechanism.
Compared to traditional NMT models, such as RNN-based and CNN-based models, the attention mechanism used in transformer can efficiently draw global dependencies between inputs and outputs, and allows significantly more parallelization.

The encoder and decoder of transformer both consist of several stacked blocks.
Each block is composed of three types of layers including (1) a multi-headed self-attention layer that takes in a set of hidden states of the previous block and weighs their their relevance to each other; (2) an optional multi-headed attention layer over the encodings to draw relevant information from the encoder outputs, which only presents in decoder blocks; and (3) a feed-forward layer that applies non-linear transformations. 
Between two consecutive layers, a residual connection and a layer normalization operation is performed. 

Model-level dual learning \cite{xia2018model} can be used to improve the translation model by leveraging the dualities of the machine translation task.
It designs a model architecture where a single model can perform both the primal task and the dual task simultaneously with the same set of parameters.
In order to apply the model-level dual learning method to the NMT transformer model, we perform some minor modifications to the transformer encoder and decoder, and construct a conditional encoder module that can perform either encoding or decoding.

\SetKwInput{KwInput}{Input}
\SetKwInput{KwOutput}{Output}

\begin{algorithm}[tb]
\DontPrintSemicolon
\caption{Model-level Dual Training of NMT Predictor}
\label{alg:dual}

\KwInput{Parallel Corpus $D$, an NMT predictor composed of two conditional encoders $Enc_A$ and $Enc_B$}
\While{not converged}
{
  sample source and target sentences $(x,y)$ from $D$ \\ 
  
  \BlankLine 

  \tcp{the primal task} 
  encode $x$ with $Enc_A$ to derive $h_x$ \\
  decode $h_x$ with $Enc_B$ to derive predictions $\tilde{y}$\\
  update the parameters of $Enc_A$ and $Enc_B$ with $loss(y,\tilde{y})$\\
  
  \BlankLine 

  \tcp{the dual task}
  encode $x$ with $Enc_B$ to derive $h_y$ \\
  decode $h_y$ with $Enc_A$ to derive predictions $\tilde{x}$\\
  update the parameters of $Enc_A$ and $Enc_B$ with $loss(x,\tilde{x})$\\
}

\end{algorithm} 

Specifically, the unified block module has the same model structure with a transformer decoder block, but functions differently in the encoding and decoding stage.
When the block works in a decoder, it performs the identical operations with a transformer decoder block.
And when the unified block performs encoding functions, instead of the encoder outputs, a tensor of zeros is passed as the $Q$ and $V$ input into its second multi-attention layer.
The attention to a tensor of zeros will result in an output of zeros as well, which is then restored back to the output of the first self-attention layer through the residual connection.
Through these operations, it is guaranteed that the unified conditional encoder will function identically with the traditional transformer encoder and decoder, with a unified structure. 
The workflow of the $i$-th unified block is shown as follows:

\begin{equation}
\label{eq:encode}
\begin{aligned}
  \mathbf{e_i} & = MultiHead(K=\mathbf{e_{i-1}}, Q=\mathbf{e_{i-1}}, V=\mathbf{e_{i-1}}), \\
  \mathbf{e_i} &= LayerNorm(\mathbf{e_i}), \\
  \mathbf{e_i} &\mathrel{+}= 
  \begin{cases}
    MultiHead(K=\mathbf{e_{i-1}}, Q=\mathbf{0}, V=\mathbf{0}), & \text{encoding}, \\
    MultiHead(K=\mathbf{e_{i-1}}, Q=\mathbf{\hat{e}}, V=\mathbf{\hat{e}}), & \text{decoding}, 
  \end{cases}\\
  \mathbf{e_i} &= LayerNorm(\mathbf{e_i}), \\
  \mathbf{e_i} &\mathrel{+}= FeedForward(\mathbf{e_i}), \\
  \mathbf{e_i} &= LayerNorm(\mathbf{e_i}),
\end{aligned}
\end{equation}
where $\mathbf{e_{i-1}}$ is the output of the previous block, and $\mathbf{\hat{e}}$ is the encoder output required in decoding processes.
By simply stacking a few blocks, we can build a unified conditional encoder that can perform exchanging tasks of encoding and decoding.

We build our NMT predictor based on the encoder-decoder architecture with two unified conditional encoders. 
By alternatively switching the two conditional encoders, we are able to train the NMT predictor on both the primal task of source-to-target translation and the dual task of target-to-source translation, with double the training data compared to the training on a single task.
Fig.\ref{fig:dual} gives an illustration of the model-level dual learning of the NMT predictor consisting of two conditional encoders.
The detailed training procedures of the NMT predictor is demonstrated in algorithm \ref{alg:dual}, where the model is trained alternatively on the primal and dual translation tasks.

\subsubsection{Mixture Models for MT}
The one-to-many mapping property of machine translation is one critical reason why it is a challenging task, as multiple valid translations exist for a given source sentence. 
The auto-regressive setting of transformer makes it difficult to search for multiple modes of solution methods, due to the fact that all uncertainty is represented in the decoder output distribution.

Commonly used decoding algorithms such as beam search allow some diversity in the decoding results. 
But they are limited to produce only minor differences in the suffixes.
Mixture models are another option to model uncertainty and generate diverse translations through different mixture components. 

Formally, $K$ expert mixture models are trained and combined, where in each mixture model, a multinomial latent variable $z \in \{1,...,K\}$ is introduced and the objective of machine translation is decomposed as 
\begin{align}
\label{eq:mix}
  p(y|x) = \sum_{z=1}^K p(y,z|x) = \sum_{z=1}^K p(z|x)p(y|z,x),
\end{align}
where the prior $p(z|x)$ and likelihood $p(y|z,x)$ are learned functions.

The posterior probability $p(z|x,y)$ can be utilized to determine the responsibility of each expert, and weigh the gradients of parameter update for each expert.
However, in this paper, we use a hard assignment of responsibilities, where each expert $z'$ is used to generate a hypotheses, and the expert giving the highest joint probability takes full responsibilities while others take none:
\begin{align}
\label{eq:res}
  r_z = 
  \begin{cases}
  1, &z=\argmax_{z'} p(y,z'|x) , \\
  0, &otherwise.\\
  \end{cases}
\end{align}
This hard assignment encourages the implication that each particular sentence should be handled only by one expert who is specialized on it, so that each expert mixture model can individually master its own translation styles.

Moreover, we set the prior distribution $p(z|x)$ to uniform in order to push all experts to generate good hypotheses for any sample and alleviate the degeneracy that only one expert gets trained well due to the ``rich get richer'' effect.
The expert models are trained using the EM algorithm \cite{mclachlan2007algorithm} and the objective function with uniform prior and hard assignments is formulated as:
\begin{align}
\label{eq:mix_obj}
  L = \mathbb{E}_{(x,y)\sim D}\big[\min_z -\log p(y|z,x)\big].
\end{align}

For the design of the expert mixture models, there are two options of either utilizing an independent decoder for each expert or using a same decoder network but with different start-of-sentence token embeddings. 
We pick the latter option because it does not overly increase the model size and the shared decoder allows even the poor experts to be trained which further mitigates the degeneracy.

\subsection{XLM Predictor}
One of the hottest topic in nature language processing is the language model pre-training, where a language model is firstly trained on a large text corpus, and fine-tuned on specific NLP tasks.
The pre-trained language models (PLMs) have achieved great success and become a must-have module in many NLP applications.
Despite the effectiveness, most PLMs are trained on monolingual corpora and only focus on English language tasks, which can hardly fulfill the needs of multi-lingual tasks such as machine translation.

Cross-lingual Language Models (XLMs) \cite{conneau2019cross} aim to build a universal cross-lingual language encoder that maps sentences of different languages into a shared embedding space.
The pre-training of an XLM model involves both monolingual and cross-lingual objectives, including masked language modeling and translation language modeling.

Masked Language Modeling (MLM) is a typical objective in language model pre-training with unsupervised monolingual corpora, in which a portion of the input tokens are randomly selected, and replaced by $[MASK]$ tokens, or random tokens, or stay unchanged. 
The altered input sentences are then passed through a transformer encoder to model the correlations between the input tokens, with the objective being to correctly recover the masked tokens based on their preceding and following contexts.

In addition to MLM, XLM models are also pre-trained on supervised cross-lingual corpora through Translation Language Modeling (TLM). 
The objective of TLM is an extension of MLM, that is also to recover the masked tokens. 
However, the inputs of TLM includes a concatenated translation sentence pair.
And the words in both the source sentence and the target sentence are randomly masked.
When predicting a masked word, the model can attend not only to surrounding words in the sentence, but also the words in the the other language, encouraging the model to align the representations of the two languages.

Through the pre-training of MLM and TLM, given a translation pair, cross-lingual language models can effectively model the cross-lingual representations and produce meaningful features for quality estimation.

\subsection{QE Feature Extraction}

The predictors trained on the parallel corpus will serve as feature extractors for the quality estimation task.
It should be noted that there exists discrepancy between the machine translation task and the quality estimation task due to the fact that unlike translation datasets, the target sentences in QE datasets are not the golden references with few translation errors.
The mistakes contained in translated sentences are necessary for the learning of the quality estimation models to distinguish between good and bad translations.
Due to this discrepancy, directly using the hidden representations outputted from the trained predictors would not be the optimal solution.
In our design, we follow the idea of previous work \cite{fan2019bilingual} to include mis-matching features and model derived features, and propose a new design of dual model features based on the model-level dual learning.

\subsubsection{Mis-matching Feature.}
The basic intuition of mis-matching features is to measure the difference between the given translation sentences and the prior knowledge of the predictors.
Specifically, for each position in the target sentence, the predictors will predict the probability $p(y_i|x)$ for each word $y_i$ in the target vocabulary.
Theoretically, a well-trained predictor is leaning to output high probabilities for correct translations in the sentence, and output low probabilities for mis-translated or missing words.

Formally, we include four dimensions in the mis-translated feature vector for each token:
\begin{equation}
\label{eq:mismatch_fea}
\begin{aligned}
  f_{mm} = \big[&logp(y_{mt}|x), logp(y_{max}|x), \\
  &logp(y_{mt}|x) - logp(y_{max}|x), \mathbf{1}_{y_{mt}=y_{max}} \big],
\end{aligned}
\end{equation}
where $logp(\cdot)$ denotes the log probability function, $y_{mt}$ is the token in the given translation, and $y_{max}$ is the token with maximum probability output from the predictor.

\subsubsection{Model derived Feature.}
Similar to QEFVs, in model derived features, the token embedding is often used in quality estimation models \cite{kim2017predictor,fan2019bilingual} to project the hidden representation $z$ derived from the predictors for each target token $y$ to the desired representations for QE task.
In our model, the hidden representation $z$ is the decoder output before the final prediction layer, derived by first encoding the source sentence $x$ with a conditional encoder $EncA$ and decoding the encoder output $h_x$ together with the target $y$ with $EncB$, 
\begin{equation}
\label{eq:model_fea}
\begin{aligned}
  h_x &= EncA.encode(x)\\
  z &= EncB.decode(h_x, y)
\end{aligned}
\end{equation}
The final feature $f_{model}=z \odot e$ is derived as the element-wise product of the hidden representation $z$ and the target embedding $e$.

\subsubsection{Dual Model Feature.}
Model-level dual learning allows the NMT predictor to be trained on both translation directions, thereby, offering the possibility of solely encoding the target information without considering the source contexts.
Based on the encoding, we propose a novel feature named Dual Model Feature.

Normally, in an NMT model, the target sentence information is only involved in the decoding stage with the source sentence encoding, to predict the next word in the target sentence while the following words are masked.
Such training setting prevents us from retrieving a good representation of the target sentence.
Fortunately, our NMT predictor is composed of conditional encoders and trained on the target-to-source translation task, meaning that the encoding of target sentences are also well trained.
The encoding of the target sentence $y$ is derived as 
\begin{align}
\label{eq:dual_fea}
  z' = EncB.encode(y).
\end{align}
And the dual model feature $f_{dual}=z'_i \odot e_i$ is derived similarly to $f_{model}$.

Compared with the model derived feature $f_{model}$ which extracts information based on source context and masked translation, our dual model feature $f_{dual}$ focuses on the internal correlations between translation words and constructs a compact representation on the target side.

\subsection{Estimator}

\begin{table*}[t]
    \centering
    \caption{Description of the WMT20 QE dataset.}
    \label{tab:datsets}
    \resizebox{17cm}{!}
    {
      \begin{tabular}{c|ccc|ccc|ccc}
        \toprule
        & \multicolumn{3}{c|}{Overall} & \multicolumn{3}{c|}{Sentence-Level} & \multicolumn{3}{c}{Word-Level}\\
        \midrule
        Dataset & Num Samples & Avg Src-Len & Avg Tgt-Len & Avg HTER & Min HTER & Max HTER & Avg BAD Ratio & Min BAD Ratio & Max BAD Ratio\\
        \midrule
        train & $7,000$ & $16.51$ & $17.15$ & $0.628$ & $0.0$ & $1.0$ & $0.305$ & $0.0$ & $0.571$\\
        dev & $1,000$ & $16.31$ & $17.06$ & $0.603$& $0.0$ & $1.0$ & $0.290$ & $0.0$ & $0.526$\\
        test & $1,000$ & $16.77$ & $17.23$ & $-$ & $-$ & $-$ & $-$ & $-$ & $-$\\
        \bottomrule
        \multicolumn{10}{l}{%
            \begin{minipage}{20cm}~\\
            \small * We report the statistics of the WMT20 QE dataset including the average sentence length of the source and target sentences, and the average, minimum, and maximum value of HTER score and the proportion of BAD tokens in all tokens.
            \end{minipage}
        }
      \end{tabular}
    }
\end{table*}

After we extract the features from the predictors, an estimator is applied to learn a transformation function that maps the features to the estimation output. 
The task of word and sentence-level quality estimation can be treated as sequential labeling and regression problems.
Therefore, we have designed our estimator based on a bidirectional GRU network connected with a few feed-forward layers.

The designed features are first concatenated together along the depth direction for both the NMT predictor and the XLM predictor.
The concatenated features are then fed to the a bi-GRU network followed by a few feed-forward layers to predict the HTER score or word-level tags for the sentence and word-level QE respectively.
In order to alleviate the overfitting problem, the weights of the Bi-GRU networks handling NMT and XLM features are shared.

\section{Experiments}

Our proposed framework is evaluated on the WMT20 English-Chinese word and sentence-level quality estimation tasks. 
We have also conducted experiments on filtering bilingual corpus based on the evaluation results to show the potential application of our proposed QE model. We will first introduce the dataset used in our evaluation and provide the implementation details along with the performance results.

\subsection{Datasets, Metrics and Baselines}

\begin{table*}[tb]
  \caption{Evaluation results of different models on WMT20 sentence-level and word-level QE tasks.}
  \label{tab:experiments}
  \begin{tabular}{c|ccc|ccc}
    \toprule
    & \multicolumn{3}{c|}{Sentence-Level} & \multicolumn{3}{c}{Word-Level} \\
    \midrule
    \textbf{Model} & \textbf{Pearson} & \textbf{MAE} & \textbf{RMSE} & \textbf{MCC} & \textbf{F1-BAD} & \textbf{F1-OK} \\
    \midrule
    NUQE \cite{martins2017pushing} & $0.4661$ & $0.1513$ & $0.1855$ & $0.4900$ & $0.6492$ & $0.7826$ \\
    PredEst \cite{kim2017predictor} & $0.5058$ & $0.1470$ & $0.1814$ & $0.5091$ & $0.6576$ & $0.8491$ \\
    Bilingual Expert \cite{fan2019bilingual} & $0.5463$ & $0.1429$ & $0.1762$ & $0.5427$ & $0.6616$ & $0.8768$ \\
    Bert-PredEst \cite{kepler2019unbabel} & $0.5676$ & $0.1449$ & $0.1790$ & $0.5501$ & $0.6883$ & $0.8226$ \\
    XLM-PredEst \cite{kepler2019unbabel} & $0.5991$ & $0.1382$ & $0.1713$ & $0.5627$ & $0.6972$ & $0.8577$ \\
    \textit{Verdi} & $\mathbf{0.6353}$ & $\mathbf{0.1342}$ & $\mathbf{0.1665}$ & $\mathbf{0.5806}$ & $\mathbf{0.7021}$ & $\mathbf{0.8785}$\\
    \midrule
    WMT20 winner (ensemble) & $0.6641$ & $0.1293$ & $0.1596$ & $0.5872$ & $\mathbf{0.7137}$ & $0.8662$\\
    \textit{Verdi} (ensemble) & $\mathbf{0.6672}$ & $\mathbf{0.1280}$ & $\mathbf{0.1588}$ & $\mathbf{0.5889}$ & $0.7006$ & $\mathbf{0.8865}$\\
    \bottomrule
    \multicolumn{7}{l}{%
    \begin{minipage}{10cm}~\\
        \small * We compare our single model of \textit{Verdi} with other single model baselines and our ensemble model with the winner of WMT20 QE competition.
    \end{minipage}
    }
  \end{tabular}
\end{table*}

In our \textit{\textit{Verdi}} framework, a parallel translation dataset is used to train the NMT predictor, and the estimator is trained on the QE dataset. For the machine translation task, we choose the CCMT-20 Chinese-English translation dataset and filter out samples with sentence length greater than 70, resulting in 8.95M data samples.
The average lengths for English and Chinese sentences are $19.66$ and $18.61$ respectively.

The QE dataset contains $7,000$ training samples, $1,000$ validation samples, and $1,000$ on-line test samples.
Ground-truth labels are not accessible for the on-line test samples, and the evaluation is performed on-line by submitting predictions to the on-line system.
For each sample, a source sentence and its translation output from a machine translation system are present.
And the labels include a sentence-level HTER score defined as the ratio of the number of edits required on the translation to its sentence length, and word-level tags for each token and gap indicating whether a token is correctly translated and if there is missing tokens in the translation sentence.

The detailed statistics of the QE dataset is summarized in table \ref{tab:datsets}. 
We can see that the overall translation quality is not satisfying with an average HTER score above $0.6$ and the BAD tokens proportion around 30\%.
Among different samples, the translation quality varies a lot.
Perfect translations are present in the dataset with $0.0$ HTER and no BAD tokens, whereas poorly translated sentences of $1.0$ HTER and with 50\% BAD tokens also exist.

We report the evaluation results with different metrics for word and sentence-level QE tasks shown as follows.
Note that the evaluation of word-level QE is done with word tags and gap tags combined together, as requested in the WMT20 QE competition.

\begin{itemize}
  \item \textbf{Pearson Correlation Coefficient} \cite{benesty2009pearson} is a statistics that measures linear correlation between two variables. It has a value between $-1$ and $+1$, where a value of $0$ means no linear correlation, and values greater than $0$ indicates positive correlations and vise versa. It is reported for the sentence-level QE task.
  \item \textbf{MAE and RMSE}, i.e., mean absolute error and root mean squared error, are two of the most common metrics used to measure accuracy of continuous variables. They are also reported for the sentence-level QE task. 
  \item \textbf{Matthews Correlation Coefficient (MCC)} \cite{matthews1975comparison} is used to measure the quality of binary classifications. Its value also ranges from $-1$ to $+1$, where $-1$ and $+1$ mean total disagreement and total agreement, and $0$ is close to random guesses. The MCC metric is reported in the word-level QE task.
  \item \textbf{F1-BAD and F1-OK} are the two F1 scores for tokens labeled with ``BAD'' and ``OK'', where the F1 score is a measure of accuracy calculated from the precision and recall values.
\end{itemize}

In the experiments, we compare our methods with the following baseline methods. 

\begin{itemize}
  \item \textbf{NUQE} \cite{martins2017pushing} uses a lookup layer containing embeddings for target words and their source-aligned words. The embeddings are concatenated and fed into two sets of feed-forward layers and a Bi-GRU layer. NUQE does not require training on parallel corpora.  
  \item \textbf{PredEst} \cite{kim2017predictor} is a OpenKiwi implementation \cite{kepler2019openkiwi} of the predictor-estimator framework, which uses a bidirectional LSTM and two unidirectional LSTMs as the predictor, and the estimator takes features produced by the predictor and makes the predictions.
  \item \textbf{XLM-PredEst and Bert-PredEst} \cite{kepler2019unbabel} combine pre-trained language models with the predictor-estimator framework through transfer learning approaches. Pre-trained XLM and Bert models are utilized respectively in these two baseline methods.
  \item \textbf{Bilingual Expert} \cite{fan2019bilingual} is a QE system proposed by \citeauthor{fan2019bilingual}. They build a conditional target language model with a bidirectional transformer and propose to extract mis-matching features from the predictor. A Bi-LSTM model is used as the estimator.
\end{itemize}

\subsection{Experimental Settings}

Our implementation is based on the Fairseq sequence modeling toolkit \cite{ott2019fairseq}.
And we use four Tesla P40 gpus to train the NMT predictor and one Tesla P40 to train the estimator.

Before we train the NMT predictor, we perform preprocessing on the CCMT20 English-Chinese translation dataset.
We first use LTP \cite{che2020n} to tokenize the Chinese sentences and perform BPE tokenization on both English and Chinese languages.
And we construct a shared vocabulary of size 70,000 containing both English and Chinese tokens.

We construct our NMT predictor based on the transformer-base architecture defined in \cite{vaswani2017attention}, with 6 encoding and decoding layers, 8 attention heads, 2,048 units for the feed-forward layers, and an embedding dimension of 512.
The mixture model technique is applied to the NMT predictor by adding 5 different start-of-sentence tokens to the vocabulary as 5 experts.
For the XLM predictor, we utilized an official released model pre-trained on both Masked Language Model and Translation Language Modeling tasks.

During the feature extraction from the NMT predictor, we extract from all the 5 expert mixture models and output a feature vector with $5 \times 512$ dimensions for both model-derived features $f_{model}$ and dual model features $f_{dual}$. Together with the 4-dimension mis-matching features, the final QE feature for each token has a dimension of $5140$.
As dual model features cannot be derived from the XLM predictor, we replicate the model derived features to fill in the position. 

The estimator consists of a bidirectional GRU with a hidden size of 512, followed by a few layers that differs in the word and sentence-level QE tasks.
In sentence-level QE, we perform a top-k operation with $k=3$ followed by max-pooling on the Bi-GRU outputs of both the NMT features and the XLM features.
The outputs are concatenated and fed into two linear layers to make the final prediction. 
MSE loss is used to update the network parameters.

In word-level QE, we first integrate the features of BPE tokens that belong to the same word into one compact representation by mean-pooling.
The processed features are then handled by the Bi-GRU network.
And different from sentence-level QE, we pass the Bi-GRU outputs directly to two FC layers with no pooling operations.
In practice, we discover that the convergence on NMT and XLM features is hard to synchronize.
Therefore, we propose to use a weighted average with weights of $0.8$ and $0.2$ to combine the NMT and XLM predictions.
Cross entropy loss is utilized to update the network.
We treat gap tag prediction as a separate task and apply the same method as predicting word tags.

As to the ensemble method, we utilize the stacking architectures \cite{wolpert1992stacked} where the outputs of various single systems are used as inputs in a second-phase regression algorithm.
To avoid overfitting, we use a five-fold cross validation, as described in \cite{martins2017pushing}.

\subsection{Main Results}

\begin{table*}[tb]
  \caption{Ablation test results of different model variants on WMT20 sentence-level and word-level QE task.}
  \label{tab:ablation}
  \begin{tabular}{c|ccc|ccc}
    \toprule
    & \multicolumn{3}{c|}{Sentence-Level} & \multicolumn{3}{c}{Word-Level} \\
    \midrule
    \textbf{Model} & \textbf{Pearson} & \textbf{MAE} & \textbf{RMSE} & \textbf{MCC} & \textbf{F1 BAD} & \textbf{F1 OK} \\
    \midrule
    NMT only & $0.5743$ & $0.1396$ & $0.1725$ & $0.5651$ & $0.6847$ & $\mathbf{0.8792}$ \\
    + XLM predictor & $0.6090$ & $0.1375$ & $0.1688$ & $0.5663$ & $0.6893$ & $0.8768$ \\
    + mixture \& dual learning & $0.6278$ & $0.1345$ & $\mathbf{0.1662}$ & $0.5789$ & $0.7006$ & $0.8783$ \\
    + dual model feature $f_{dual}$ & $\mathbf{0.6353}$ & $\mathbf{0.1342}$ & $0.1665$ & $\mathbf{0.5806}$ & $\mathbf{0.7021}$ & $0.8785$\\
    \bottomrule
  \end{tabular}
  \vspace{0mm}
\end{table*}

Table \ref{tab:experiments} first compares the performance of our proposed single model with the mentioned baseline methods, and then compares our ensemble model with the winner of WMT20 QE competition on sentence and word levels, in terms of different evaluation metrics. 
We can see that our single method achieves the best performance on both word and sentence-level QE tasks and significantly outperforms other baseline algorithms. 
On the sentence-level QE dataset, given source sentences and their corresponding translations, the Pearson correlation coefficient, MAE, and RMSE of our single model result are $0.635$, $0.134$, and $0.167$ respectively, while the best result of previous methods are $0.599$, $0.138$, and $0.171$ respectively.
Similarly, our method also achieves the best results among all methods in word-level QE task.
We also compare our ensemble result to the winning system of WMT20 QE competition and show that our ensemble model is able to beat them in both word and sentence-level quality estimation.

The reason of our success is attributed to the different strategies applied in our framework to enhance the feature extraction ability.
First, we utilized an XLM predictor together with an NMT predictor to improve the feature representations.
Second, we use model-level dual learning and mixture model technique to train the NMT predictor to produce more diverse and robust QE features.
Third, taking the advantage of model-level dual learning, we propose a novel dual model feature to model the internal correlations and construct a compact representation on the translated sentence, which further improves the QE features.
By incorporating these strategies and modules, our \textit{Verdi} framework is able to outperform other method by a significant margin.

\subsection{Analysis}

We evaluate the impact of different modules and methods by performing an ablation test for both word and sentence-level QE on WMT20 datasets. 
Table \ref{tab:ablation} lists the performances of our model variants with different modules and methods included.

We incrementally accommodate different modules and methods to a base model which is a simple model using predictor-estimator architecture with a single transformer NMT predictor, with mis-matching features and model derived features incorporated.
The base model produces a result of $0.574$ Pearson on sentence-level QE and $0.565$ MCC on word-level QE.
Based on it, we include the second XLM predictor and achieves $0.609$ and $0.566$ on sentence and word level respectively, which evidently proves the effectiveness of utilizing multiple predictors as feature extractors.
Furthermore, we apply model-level dual learning and mixture model technique to the NMT predictor. 
This model variant further produces a performance of $0.628$ Pearson and $0.579$ MCC on the two tasks, and the significant improvement can be attributed to the model's capability of generating better representations by exploiting the translation dualities and producing diverse translation outputs.
Finally, we incorporate our designed dual model features to the model, and achieve the best performance of $0.635$ Pearson score and $0.581$ MCC score, showing that constructing a more compact representation of the translation sentence is helpful to the model performance.

\subsection{Applied to Bilingual Corpus Filtering}
We also perform an experiment on bilingual corpus filtering by applying our QE model to evaluate the quality of samples in the parallel corpus.
Specifically, we run the sentence-level evaluation on 6 million samples from the CCMT20 Chinese-English dataset with a trained QE model of our framework.
We sort the 6 million samples according to their predicted HTER scores, and filter out 1 million samples with the highest error rates, resulting in a QE filtered dataset with 5 million translation pairs.

\begin{table}[tb]
  \caption{Evaluation results of the bilingual corpus filtering on CCMT20 English-Chinese dataset.}
  \label{tab:filter}
  \begin{tabular}{c|cc}
    \toprule
    \textbf{Dataset} & \textbf{Num Samples} & \textbf{BLEU}\\
    \midrule
    Unfiltered & $6,000,000$ & $20.39$\\
    QE filtered & $5,000,000$ & $\mathbf{20.48}$\\
    \bottomrule
    \multicolumn{3}{l}{%
    \begin{minipage}{5cm}~\\
        \small * We report the average of the BLEU scores measured on the best three checkpoints.
    \end{minipage}
    }
  \end{tabular}
\end{table}

Table \ref{tab:filter} compares the performance of transformer NMT model trained on the unfiltered 6 million samples and the filtered 5 million samples.
We can see that even though the filtered dataset contains fewer samples, the model trained on it produces a better performance with a BLEU score of $20.48$ compared to $20.39$ given by the model trained on the full 6 million data.
This result proves that the QE evaluation result is a good indication of the translation quality and can be potentially helpful to bilingual corpus filtering.
By removing the low quality samples, we are able to reduce the training efforts, and at the same time, improve the performance of the NMT system, which from another aspect proves the effectiveness of our QE framework.

\section{Related Work}

In this section, we review related work on quality estimation and dual learning.

The traditional studies on QE are mostly based on manual feature engineering that investigates the useful features to learn a shallow regressor/classifier to evaluate the quality of a given translation pair.
Different features have been investigated including 
linguistic features \cite{felice2012linguistic}, statistical baseline features \cite{specia-etal-2013-quest}, topic model features \cite{rubino2013topic}, and pseudo-reference and back-translation features \cite{kozlova2016ysda}, possibly with feature selection strategies such as principal component analysis \cite{gonzalez2012prhlt}.
However, the manually designed features require expert knowledge and are hard to transfer between datasets and models.

Deep learning has also been applied to quality estimation.
Context window-based approaches are widely used in many deep learning QE studies, where bilingual context windows are derived for both source and target sentences with the center words in both languages properly aligned. 
\citeauthor{kreutzer2015quality} \cite{kreutzer2015quality} proposes to adopt an FNN architecture on the context windows and combine it with a linear model using handcrafted QE features.
Recurrent networks and convolutional networks on context windows are also introduced \cite{patel2016translation,martins2016unbabel,martins2017pushing}.

Recently, the predictor-estimator framework \cite{kim2017predictor} is proposed to evaluate the translation quality without the need of word alignments.
It consists of a word prediction RNN trained on parallel corpora that extracts representative features from the source sentence and its translation, and a neural quality estimation model that makes predictions based on the extracted features.
\citeauthor{fan2019bilingual} \cite{fan2019bilingual} proposes to use a bidirectional transformer language model to produce the joint representation and designs features that measure the degree of mis-matching.
Pre-trained language models such as BERT \cite{devlin2018bert}, are also introduced into the predictor-estimator framework as the feature extractor.

Different from the existing QE methods, our work utilizes an transformer predictor together with an XLM predictor, and incorporates model-level dual learning to enhance the feature extraction ability.

Dual learning is a technique that leverages the symmetric structure of some learning problems such as machine translation. 
\citeauthor{he2016dual} \cite{he2016dual} focuses on the unsupervised setting and proposes to assist the learning of a translation model by exploiting the distortion between a sentence and its back-translation.
Dual supervised learning \cite{xia2017dual} conducts joint learning from labeled data by adding an additional probabilistic constraint.
Model-level dual learning \cite{xia2018model}, from another aspect, exploits the duality of tasks in model level.
It designs an architecture for primal and dual tasks, and ties the model parameters playing similar roles.

\section{Conclusion and Future Work}

In this paper, we present our \textit{Verdi} framework to solve the quality estimation problem.
Our predictor module includes a pre-trained XLM model and an NMT model that is trained through model-level dual learning and mixture models technique. 
Besides, we also designed a novel feature to construct a better representation on the target side.
Experiments conducted on WMT20 QE competition demonstrate that our system yields a better performance than other baselines and beats the winning system in the competition.
We have also shown a potential application of employing our quality estimation method on bilingual corpus filtering task and demonstrated its effectiveness.

For future work, we plan to investigate the possibility of extending our framework to automatic post-editing tasks, in which the model does not only detect the potential mistakes, but performs automatic corrections to improve the translation qualities.

\bibliographystyle{ACM-Reference-Format}
\bibliography{main}

\end{document}